\documentclass[twocolumn,letterpaper]{IEEEAerospaceCLS}  

\usepackage[]{graphicx}    
\usepackage[colorlinks=true, urlcolor=blue, linkcolor=black, citecolor=black]{hyperref} 
\newcommand{\ignore}[1]{}  
\usepackage{multirow}
\usepackage{booktabs}
\usepackage{amsmath}
\usepackage{amssymb}
\usepackage{subfig}
\usepackage{enumitem}
\usepackage{siunitx}
\usepackage{multicol}

\begin{document}
\title{Visual SLAM with DEM Anchoring for Lunar Surface Navigation}

\author{%
Adam Dai, Guillem Casadesus Vila, Grace Gao\\ 
Stanford University\\
William F. Durand Building\\
496 Lomita Mall\\
Stanford, CA 94305\\
\{addai, guillemc, gracegao\}@stanford.edu 
\thanks{\footnotesize 979-8-3315-7360-7/26/$\$31.00$ \copyright2026 IEEE}              
}

\maketitle

\thispagestyle{plain}
\pagestyle{plain}

\maketitle

\thispagestyle{plain}
\pagestyle{plain}

\begin{abstract}
Future lunar missions will require autonomous rovers capable of traversing tens of kilometers across challenging terrain while maintaining accurate localization and producing globally consistent maps. 
However, the absence of global positioning systems, extreme illumination, and low-texture regolith make long-range navigation on the Moon particularly difficult, as visual-inertial odometry pipelines accumulate drift over extended traverses. 
To address this challenge, we present a stereo visual simultaneous localization and mapping (SLAM) system that integrates learned feature detection and matching with global constraints from digital elevation models (DEMs). 
Our front-end employs learning-based feature extraction and matching to achieve robustness to illumination extremes and repetitive terrain, while the back-end incorporates DEM-derived height and surface-normal factors into a pose graph, providing absolute surface constraints that mitigate long-term drift. 
We validate our approach using both simulated lunar traverse data generated in Unreal Engine and real Moon/Mars analog data collected from Mt. Etna. 
Results demonstrate that DEM anchoring consistently reduces absolute trajectory error compared to baseline SLAM methods, lowering drift in long-range navigation even in repetitive or visually aliased terrain.
\end{abstract}

\tableofcontents

\section{Introduction}
\label{sec:intro}

Long-range autonomy is a key enabler for expanding future missions to the Moon and other planetary surfaces. 
Proposed missions such as the Endurance rover highlight the importance of sustained scientific exploration, with traverses on the order of 2,000 km required to achieve their objectives. 
Beyond single-mission demonstrations, current plans for a sustained lunar presence will demand robust and scalable autonomy to support logistics, transport, and in-situ resource utilization (ISRU). 
In many scenarios, long-range rover operations will be critical, as landing sites may be located far from permanent bases or science targets~\cite{merancy2024moon}. 
These drivers make reliable, drift-free navigation across tens of kilometers a central challenge for future lunar surface exploration.

However, long-range lunar navigation poses unique challenges. 
Unlike Earth, there is no dedicated lunar navigation satellite system, eliminating the possibility of consistent, absolute position fixes. 
While future lunar navigation satellite constellations are planned, they remain years from deployment.
Alternative sensors offer partial solutions but face their own limitations: LiDAR provides rich geometry but is power-intensive; inertial measurement units (IMUs) and wheel odometry can complement vision-based approaches but provide no standalone ability for obstacle detection or map generation. 
Given these constraints, vision, and stereo vision in particular, has emerged as a cornerstone of planetary rover navigation, as evidenced by its extensive heritage on Mars missions from Spirit and Opportunity through Perseverance, and its adoption in planned lunar missions including NASA's VIPER and JPL's CADRE missions.

Despite this proven track record, visual navigation pipelines accumulate drift over kilometer-scale traverses, a problem exacerbated by the lunar surface environment. 
Repetitive, low-texture regolith degrades feature tracking, and extreme illumination with long shadows and saturated highlights disrupts visual matching. 
Many planetary traverses follow long, straight paths across uniform terrain with few opportunities to revisit viewpoints, precluding loop closures that correct drift in traditional SLAM.
These factors make vision-only approaches insufficient for long-range navigation.
To address this challenge, global maps and priors—such as orbital digital elevation models (DEMs)—offer a promising avenue for providing absolute surface constraints that can anchor local odometry estimates and recover global consistency. 

\subsection{Contributions}

In this paper, we make the following contributions:
\begin{itemize}[leftmargin=1em]
    \item We present a complete stereo SLAM system tailored to long-range lunar navigation, integrating learned feature detection and matching for robustness to low-texture imagery and illumination extremes. 
    \item We introduce DEM anchoring factors that constrain the SLAM pose graph, improving long-range consistency.
    \item We validate our approach on both simulated and real-world lunar datasets, and demonstrate improved localization accuracy and global consistency.
\end{itemize}

\subsection{Paper Organization}

The remainder of the paper is organized as follows: Section~\ref{sec:related_work} reviews related work, Section~\ref{sec:vslam} describes our visual SLAM pipeline, Section~\ref{sec:dem_anchoring} introduces DEM anchoring, Section~\ref{sec:datasets} outlines the datasets and simulation environments used, Section~\ref{sec:results} presents results, and Section~\ref{sec:conclusion} concludes the paper.

\section{Related Work}
\label{sec:related_work}

Visual navigation for planetary rovers has been studied extensively, both in space missions and in academic research. We group prior work into three categories: (1) visual navigation on planetary surfaces, (2) strategies for long-range consistency, and (3) state-of-the-art SLAM methods in robotics and computer vision.



\subsection{Planetary Surface Visual Navigation}

Visual navigation on planetary surfaces has a rich heritage beginning with the Mars Exploration Rovers. Spirit and Opportunity pioneered the use of visual odometry (VO) to monitor slip and improve localization accuracy in low-texture terrains~\cite{maimone2007two}. Curiosity later adopted VO during traverse operations, while Perseverance introduced upgraded cameras, onboard processors, and the AutoNav system, achieving the longest autonomous single-sol drive of 347.7 \si{m}~\cite{verma2023autonomous}. 

On the Moon, autonomy is less mature but increasingly important. The Apollo Lunar Roving Vehicles were manually driven by astronauts, while recent and planned missions emphasize onboard navigation. China’s Yutu-2 rover (Chang’e-4) has demonstrated semi-autonomous capabilities, and NASA’s upcoming VIPER rover will carry out prospecting operations at the lunar South Pole. JPL’s CADRE (Cooperative Autonomous Distributed Robotic Exploration) technology demonstration aims to validate multi-agent autonomy on the lunar surface, while the proposed Endurance mission highlights the need for long-range autonomous traverse capability. 

Outside of missions, numerous research efforts have studied VO and SLAM in planetary analog environments. Field campaigns on Devon Island, Mount Etna, and the Moroccan desert have produced benchmark datasets for rover navigation in Mars- and Moon-like terrain~\cite{furgale2012devon,vayugundla2018datasets,meyer2021madmax}. These datasets expose the challenges of unstructured, low-texture, and aliased environments, providing a foundation for evaluating new SLAM approaches.

\subsection{Long-Range Lunar Navigation}

While VO provides locally accurate trajectories, drift accumulates rapidly over kilometer-scale traverses, motivating methods that incorporate absolute or global cues. 
Several strategies have been proposed to address this long-range challenge.
Crater-based localization, as in LunarNav~\cite{daftry2023lunarnav}, exploits distinctive topographic features visible from orbit. 
ShadowNav~\cite{atha2024shadownav} uses active illumination to detect crater edges even in total darkness, providing navigation cues in permanently shadowed regions. 
Skyline or horizon-based matching aligns rover imagery with orbital maps to provide coarse global localization~\cite{nefian2014planetary}. 
Digital elevation models (DEMs) derived from orbital have also been integrated into SLAM pipelines, either as global altitude constraints~\cite{melman2022lcns} or as priors for LiDAR-inertial SLAM~\cite{zhang2024lidar}. 
These approaches show promise for extending rover navigation to tens of kilometers, but rely on sparse, noisy, or incomplete global information. A key open problem is how to robustly fuse these absolute cues with local perception to achieve globally consistent and drift-free long-range navigation.

\subsection{SLAM in Robotics and Computer Vision}

In terrestrial robotics, a large body of work has advanced SLAM algorithms that are now widely used as baselines. 
Classical systems such as ORB-SLAM3~\cite{campos2021orbslam3} and VINS-Mono~\cite{qin2018vins} combine feature-based visual odometry with loop closure detection to achieve globally consistent mapping. 
While highly effective in structured environments, these approaches degrade in planetary conditions due to repetitive textures, low feature density, and extreme illumination variation.

Recent trends have shifted toward learning-based features and matchers for improved robustness to appearance and viewpoint changes. SuperPoint~\cite{detone_superpoint_2018-1} and R2D2~\cite{revaud2019r2d2} learn features that are more repeatable under challenging imaging conditions, while LightGlue~\cite{lindenberger2023lightglue} leverages transformers for adaptive feature matching. Lightweight SLAM systems such as LightSLAM~\cite{zhao2025light} combine LightGlue matching with efficient backends, outperforming traditional pipelines on degraded imagery. 

For planetary navigation, learned features and matchers have the potential to offer robustness to lighting extremes and repetitive textures, addressing key limitations of classical pipelines. 
Combining such robust visual front-ends with DEM-based global anchoring enables both short-term resilience and long-term consistency in lunar environments. 

\section{Visual SLAM}
\label{sec:vslam}

Our SLAM pipeline builds on our prior work for the Lunar Autonomy Challenge, which demonstrated robust short-range autonomy under simulated lunar conditions~\cite{Dai2025LAC}. 
We assume access only to calibrated stereo cameras with known intrinsic and extrinsic parameters.
The system follows a standard frontend/backend architecture: the frontend extracts features and estimates stereo visual odometry, while the backend maintains a pose graph that enforces local consistency and incorporates loop closures for global optimization. 
We structure the system to run in real-time, ingesting camera images at each timestep and incrementally building and optimizing the factor graph.
This system provides the foundation for the DEM anchoring extensions presented in Section~\ref{sec:dem_anchoring}.

\subsection{Frontend}

The frontend employs learning-based features to improve robustness to the illumination extremes and low-texture terrain characteristic of the lunar surface. 
For each left and right image, keypoints are detected with SuperPoint~\cite{detone_superpoint_2018-1}, and correspondences are established using LightGlue~\cite{lindenberger2023lightglue}, a transformer-based matcher that adapts to the distribution of features in each image. 
This combination has been shown to outperform traditional handcrafted descriptors over extreme viewpoint and lighting changes.
We use the official pretrained weights released with the respective methods with no additional fine-tuning.

Relative motion is estimated from stereo image pairs. Given matched keypoints $\{\mathbf{x}_i^l, \mathbf{x}_i^r\}$ in the left and right cameras, we triangulate 3D landmarks $\mathbf{X}_i$ and solve a Perspective-$n$-Point (PnP) problem to recover the relative pose between consecutive frames. 
Between frames $i$ and $j$, the resulting motion estimate $\hat{\mathbf{T}}_{ij} \in SE(3)$ is used to generate odometric constraints.

\begin{figure}[ht]
    \centering
    \includegraphics[width=\linewidth]{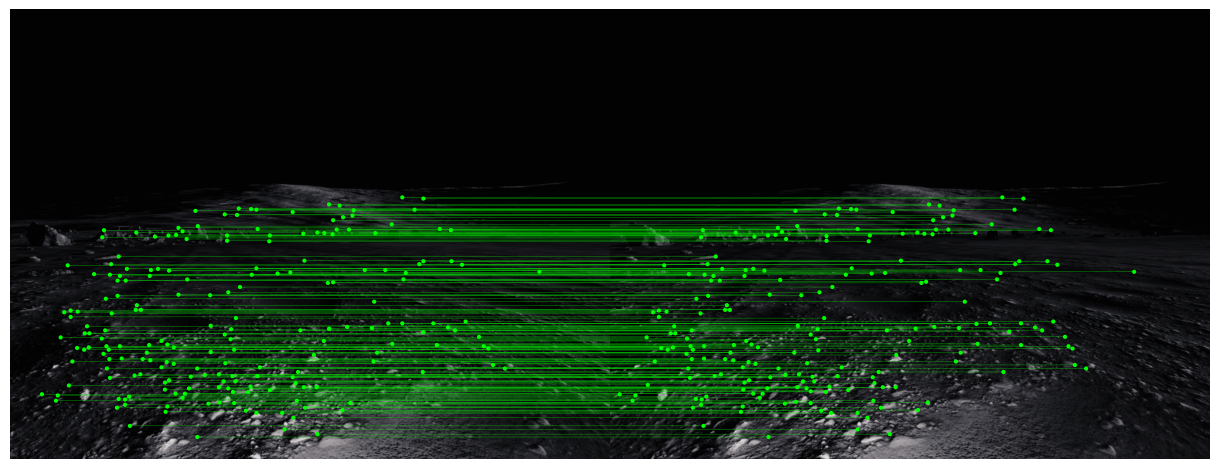}
    \caption{LightGlue feature matching between left and right stereo images from our Unreal Engine lunar simulation. Features are detected and matched reliably despite challenging illumination conditions. A maximum of 200 matches are plotted to reduce clutter.} 
    \label{fig:feat_matching}
\end{figure}

\subsection{Backend}

The back-end maintains a pose graph, where each node represents a camera pose $\mathbf{T}_i \in SE(3)$, and edges represent relative pose constraints. For an odometric edge between nodes $i$ and $j$, the residual is
\begin{equation}
\mathbf{r}_{ij} = \mathrm{Log}\left( \hat{\mathbf{T}}_{ij}^{-1} \, \mathbf{T}_i^{-1}\mathbf{T}_j \right),
\end{equation}
where $\mathrm{Log}(\cdot)$ maps from $SE(3)$ to its Lie algebra. 
Loop closure constraints are added by matching features between non-consecutive frames and verifying geometric consistency. 
Upon loop closure, the pose graph is optimized as a nonlinear least squares through Levenberg-Marquardt with GTSAM~\cite{gtsam}.

This baseline system achieves reliable and drift-minimized navigation over short traverses, as demonstrated in the LAC setting. 
However, in the absence of absolute references, accumulated error remains a limiting factor for long-range missions spanning tens of kilometers. 
To address this limitation, we extend the pose graph with DEM anchoring factors that provide global surface constraints, improving consistency over extended traverses. 
These extensions are described in the following section.

\section{DEM Anchoring}
\label{sec:dem_anchoring}

To mitigate long-term drift from visual odometry, we augment the pose graph with anchoring factors derived from a reference digital elevation model (DEM). 
The DEM provides a globally consistent surface representation, which we treat as a prior map against which rover pose estimates can be constrained.

\subsection{Digital Elevation Model}

Orbital DEMs are typically provided in a global reference frame (e.g., selenographic coordinates). 
We assume the DEM has been transformed to a local East-North-Up (ENU) frame centered on the mission area. 
The DEM is represented as a regular raster grid $h: \mathbb{R}^2 \to \mathbb{R}$, where $h(x,y)$ denotes the elevation at horizontal position $(x,y)$. 
For arbitrary query positions, we use bilinear interpolation between the four nearest grid cells. 
We precompute surface gradients $\nabla h(x,y)$ across the grid, which are needed for the surface normal constraints described below.

We assume that rover poses $\mathbf{T}_i \in SE(3)$ are expressed in the same local frame as the DEM, enabling direct comparison between estimated rover heights and DEM-derived surface values.
The representation allows us to construct two complementary constraints: (i) a height consistency factor that enforces agreement between estimated and DEM heights, and (ii) a surface normal factor that aligns the estimated local ground orientation with DEM gradients.

\subsection{Height Factor}

Let $\mathbf{t}_i = [x_i, y_i, z_i]^\top$ denote the translation of pose $\mathbf{T}_i$. 
Using the rover’s horizontal position $(x_i, y_i)$, we query the DEM height $\hat{z}_i = h(x_i,y_i)$ using bilinear interpolation. 
The height factor penalizes the discrepancy between the estimated rover height $z_i$ and the DEM surface height:
\begin{equation}
r^{\text{height}}_i = z_i - \hat{z}_i,
\end{equation}
with residual covariance $\sigma_z^2$ determined by DEM vertical error estimates. 
This factor encourages the estimated trajectory to remain consistent with the absolute surface elevation, constraining long-term drift in the vertical direction.


\subsection{Surface Normal Factor}

Beyond absolute height, DEM gradients encode local surface orientation. 
The DEM slope at $(x_i,y_i)$ is given by
\begin{equation}
\nabla h(x,y) = \left[ \frac{\partial h}{\partial x}, \frac{\partial h}{\partial y} \right]^\top,
\end{equation}
and the corresponding DEM surface normal is
\begin{equation}
\mathbf{n}_{\text{DEM}} = \frac{1}{\sqrt{\|\nabla h\|^2 + 1}} \; \begin{bmatrix} -\frac{\partial h}{\partial x} \\ -\frac{\partial h}{\partial y} \\ 1 \end{bmatrix}.
\end{equation}

From the estimated pose $\mathbf{T}_i$, we extract the rover’s local ground normal $\mathbf{n}_i$ (e.g., the $z$-axis of the body frame). 
The surface normal factor enforces alignment between these directions:
\begin{equation}
r^{\text{normal}}_i = \mathrm{Log}\left( \mathbf{R}(\mathbf{n}_{\text{DEM}})^{-1} \, \mathbf{R}(\mathbf{n}_i) \right),
\end{equation}
where $\mathbf{R}(\mathbf{n}) \in SO(3)$ denotes the minimal rotation aligning the world $z$-axis with normal $\mathbf{n}$. This factor penalizes orientation drift, particularly in pitch and roll, by constraining the trajectory to follow the DEM surface geometry.



Together, the height and surface normal factors provide complementary global information: the height factor anchors vertical position, while the normal factor constrains orientation drift. These constraints are added to the pose graph alongside relative odometry and loop closure edges, producing a globally anchored trajectory. 
To maintain computational efficiency over long trajectories, we add DEM factors periodically rather than at every pose—typically every 10 to 100 poses depending on traverse length and terrain variability.

\section{Datasets and Simulation}
\label{sec:datasets}

\begin{figure*}[t]
    \centering
    \begin{tabular}{ccc}
        \includegraphics[width=0.3\linewidth]{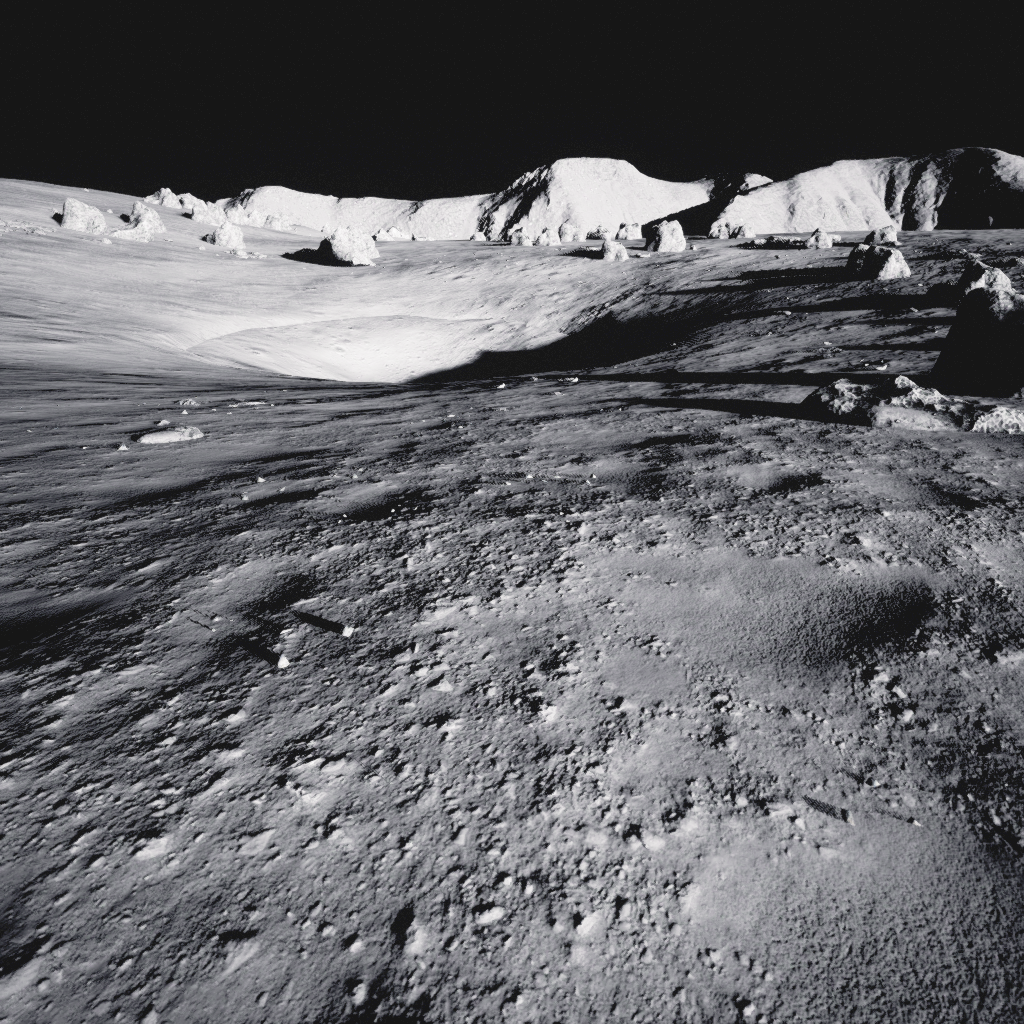} &
        \includegraphics[width=0.3\linewidth]{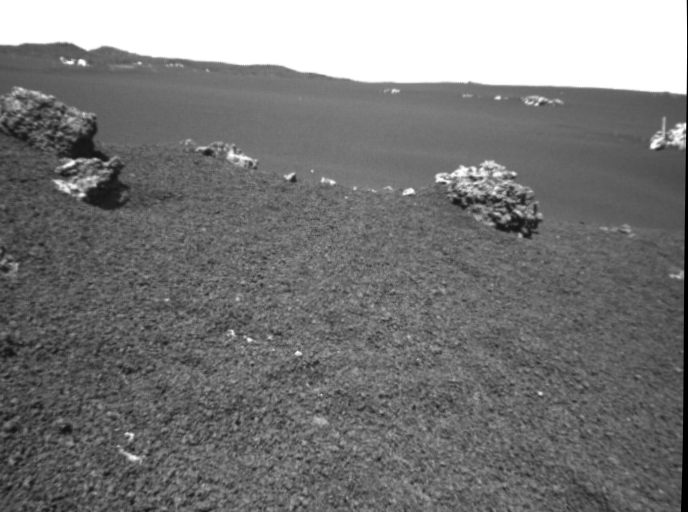} &
        \includegraphics[width=0.3\linewidth]{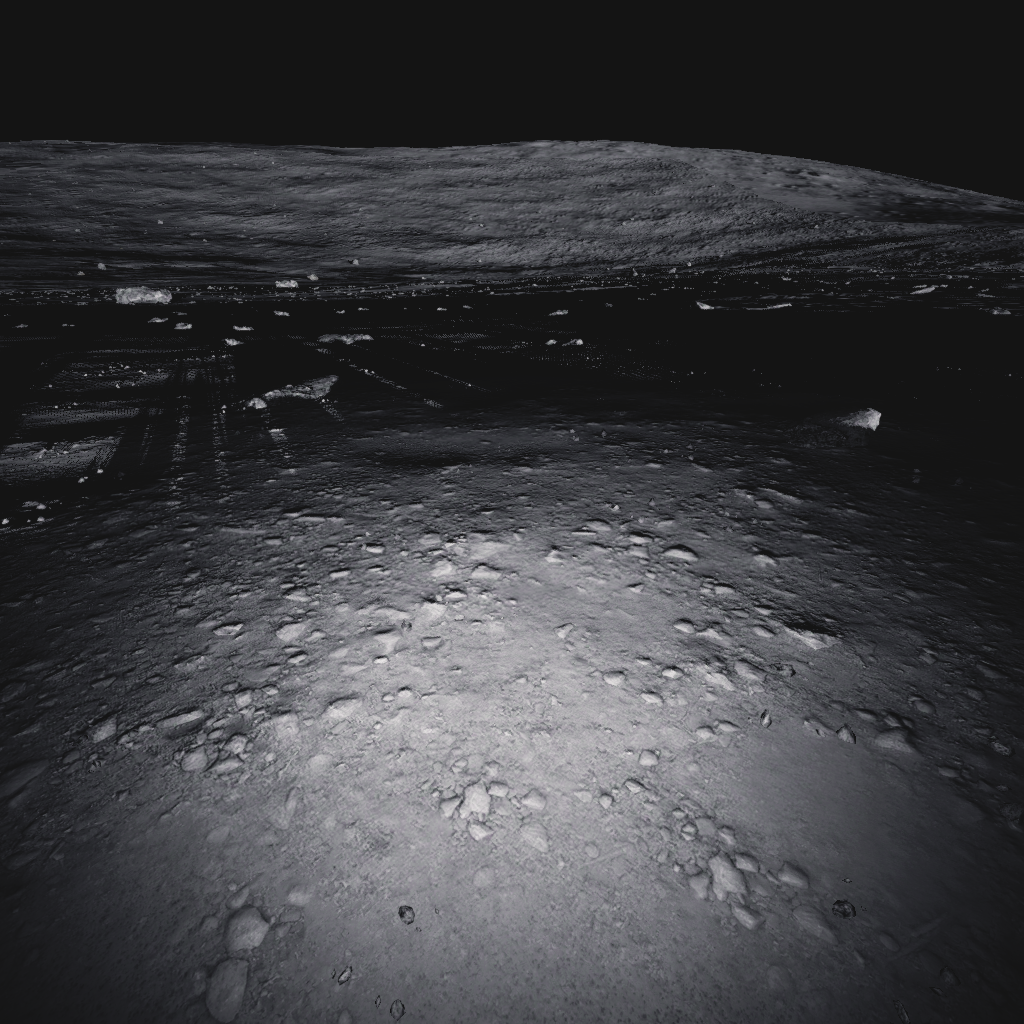} \\
        \includegraphics[width=0.3\linewidth]{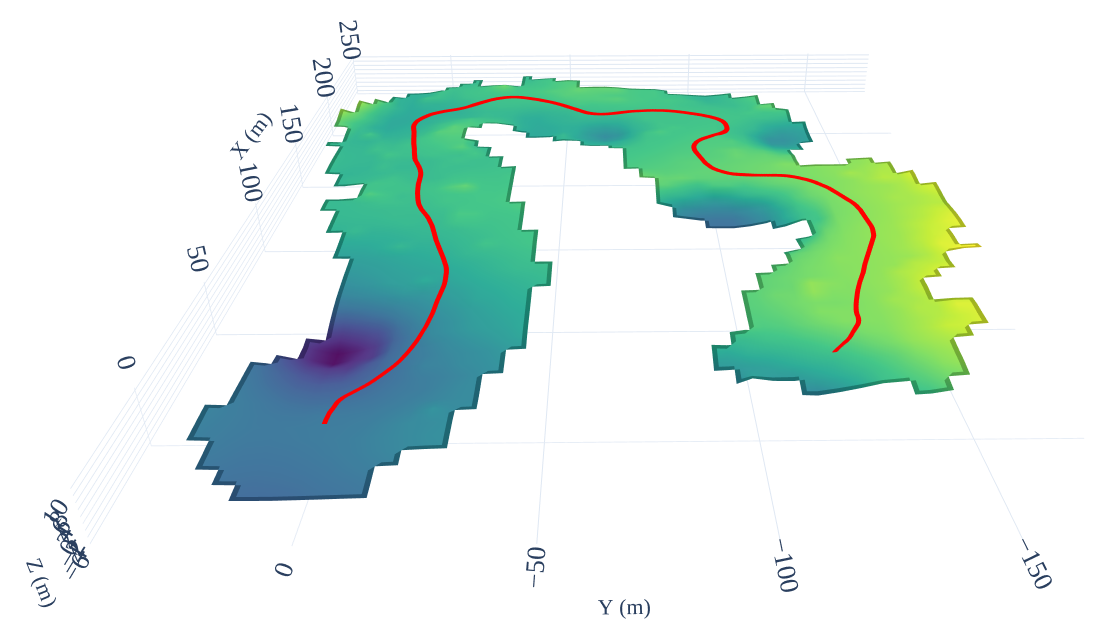} &
        \includegraphics[width=0.3\linewidth]{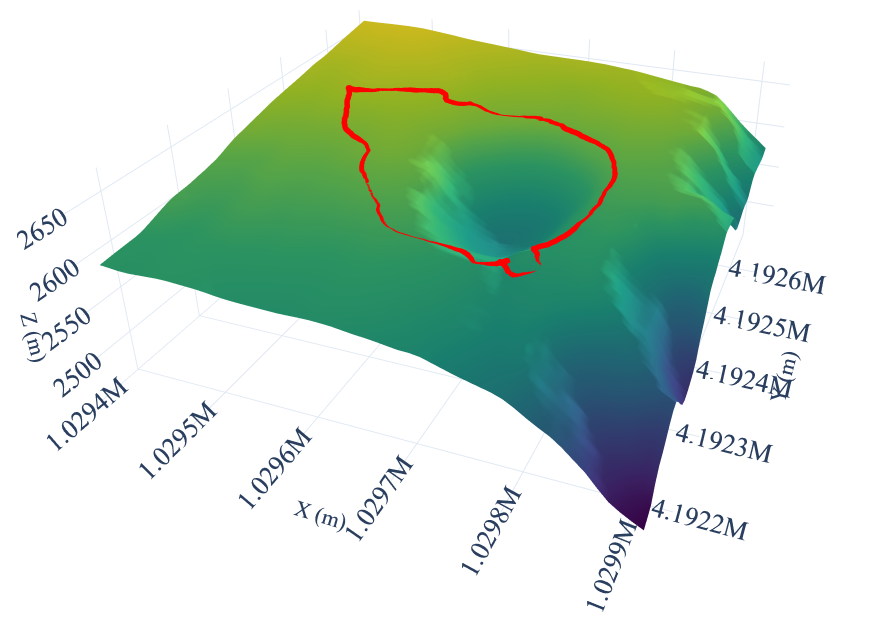} &
        \includegraphics[width=0.3\linewidth]{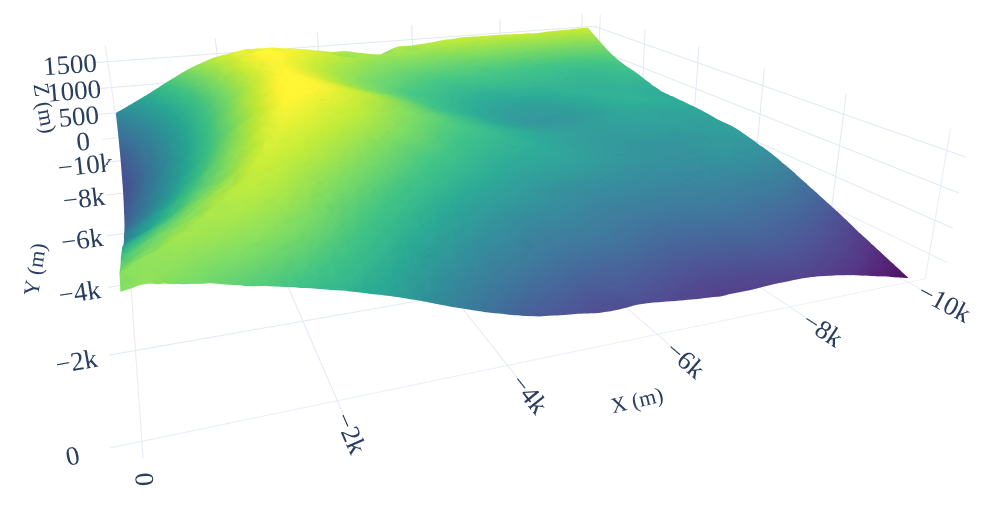} \\
    \end{tabular}
    \caption{Example data from each dataset. Top: sample rover imagery. Bottom: corresponding digital elevation models (DEMs). From left to right: LuSNAR, S3LI, Unreal Engine simulation.}
    \label{fig:datasets}
\end{figure*}

To evaluate our approach, we draw on three complementary sources of data: (i) the LuSNAR dataset, which provides diverse synthetic lunar environments; (ii) the S3LI dataset, which offers real-world stereo, LiDAR, and inertial data recorded in a lunar analog site; and (iii) a custom Unreal Engine simulator built on top of high-resolution LOLA digital elevation models of the lunar South Pole. Together, these datasets enable us to test our system under both controlled conditions and natural, unstructured environments.

\subsection{LuSNAR}

The LuSNAR dataset~\cite{liu2024lusnar} is a synthetic multi-sensor benchmark designed for lunar exploration tasks. It provides stereo imagery, dense depth maps, semantic labels, and LiDAR point clouds across nine scenes with varying levels of terrain relief and object density, and trajectory lengths ranging from 200 to 500 \si{m}. While LuSNAR offers high-fidelity visual and geometric data, it does not include global DEMs for its environments. To enable DEM anchoring, we manually construct local DEMs by accumulating LiDAR point clouds along each trajectory and projecting them into a 2.5D grid representation. This allows us to test our method under controlled synthetic conditions while still incorporating DEM constraints.

\subsection{S3LI}

The DLR Planetary Stereo, Solid-State LiDAR, Inertial (S3LI) dataset~\cite{giubilato2022challenges} was collected on Mount Etna, Sicily, a planetary analog site with visual and structural characteristics similar to the Moon. The dataset was recorded using a hand-held sensor suite that replicates a rover’s vantage point and includes a stereo monochrome camera, a solid-state LiDAR, and an IMU. Accurate ground truth is provided via differential GNSS (D-GNSS). Seven sequences are available, spanning 300 \si{m} to over 1 \si{km}. We focus on the \texttt{s3li\_crater} sequence, the longest at 1.03~\si{km}, which consists of a loop around the rim of the Cisternazza crater. 
This sequence provides a particularly challenging test for long-range SLAM, with severe visual aliasing, minimal landmarks, and extended traverse length.

For the corresponding DEM, we use a high-resolution Digital Surface Model (DSM) of Mt. Etna~\cite{palaseanu2020digital} derived from Pleiades satellite data using the NASA Ames Stereo Pipeline (ASP), with 2 \si{m} spatial resolution and vertical RMSE of 0.78 \si{m}.
To align the ground truth positions (provided in WGS84 lat-lon-alt) with the DEM, we convert both to UTM coordinates and then shift to a common local frame.

\subsection{Unreal Engine Simulation}

\begin{figure}[t]
    \centering
    \includegraphics[width=0.8\linewidth]{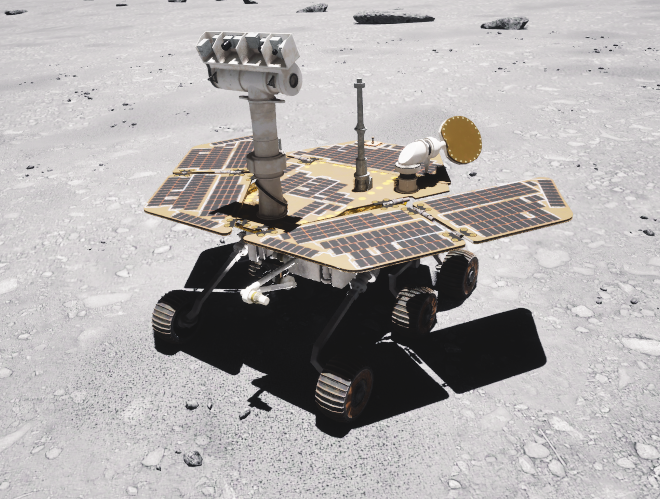}
    \caption{Opportunity rover model~\protect\cite{opportunity_rover_asset} used in our custom Unreal Engine simulation.}
    \label{fig:unreal_rover}
\end{figure}

To complement these datasets, we develop a custom simulator in Unreal Engine 5 that supports DEM interfacing, photorealistic rendering, and precise control over rover and sensor configurations. 
Our simulator is developed from LuPNT~\cite{vila2025lupnt}, using custom APIs to connect to Unreal Engine via Python and C++.
The simulator enables rendering from multiple cameras; in this work, we use a front-facing stereo pair to replicate our hardware setup. 
The virtual terrain is built from Lunar Orbiter Laser Altimeter (LOLA) DEMs of the South Pole~\cite{barker2021lola}, including both a high-resolution 10~\unit{km} $\times$ 10~\unit{km} patch at 5 \unit{m}/pixel for local traverses and a larger 120~\unit{km} $\times$ 120~\unit{km} region at 30 \unit{km}/pixel for far-field context. 
On top of the LOLA DEM landscape, we also add procedurally generated terrain deformations and rocks of varying size which reflect small-scale irregularities which may not be observed from orbit.
We render images at a resolution of 1024 $\times$ 1024 pixels with a $90^\circ$ field of view. 
This environment allows us to evaluate our system in realistic lunar lighting conditions while ensuring complete ground-truth availability for analysis.
\section{Results}
\label{sec:results}

\begin{figure*}[t]
    \centering
    \includegraphics[width=0.8\linewidth]{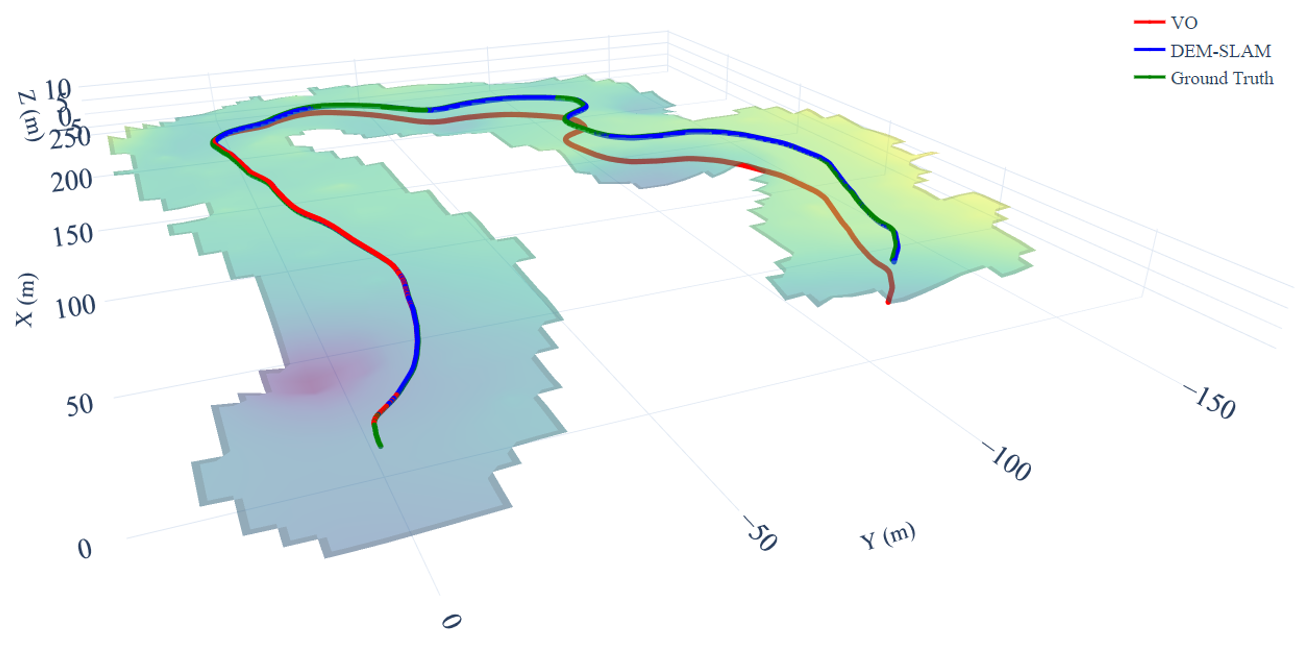}
    \caption{Trajectories from LuSNAR scene 9. Ground-truth trajectory is shown in green, estimated VO trajectory in red, and DEM-anchored SLAM in blue. DEM anchoring successfully constrains the SLAM trajectory to the surface and mitigates vertical drift.}
    \label{fig:lusnar_trajs}
\end{figure*}

\begin{table*}[ht!]
  \centering
  \caption{\bfseries RMSE comparison of different SLAM methods across nine LuSNAR scenes. Best values are in \textbf{bold}.}
  \label{tab:lusnar_results}
  \renewcommand{\arraystretch}{1.3}
  \begin{tabular}{|c|cc|cc|cc|}
    \hline
    \multicolumn{1}{|c|}{\multirow{2}{*}{\textbf{Scene}}} &
    \multicolumn{2}{|c|}{\textbf{VO}} &
    \multicolumn{2}{|c|}{\textbf{DEM-SLAM}} &
    \multicolumn{2}{|c|}{\textbf{ORB-SLAM3}} \\
    \cmidrule(lr){2-7}
    \multicolumn{1}{|c}{} &
    \multicolumn{1}{|c}{\textbf{RPE [m] $\downarrow$}} &
    \multicolumn{1}{c|}{\textbf{ATE* $\downarrow$}} &
    \multicolumn{1}{|c}{\textbf{RPE [m] $\downarrow$}} &
    \multicolumn{1}{c|}{\textbf{ATE* $\downarrow$}} &
    \multicolumn{1}{c}{\textbf{RPE [m] $\downarrow$}} &
    \multicolumn{1}{c|}{\textbf{ATE* $\downarrow$}} \\
    \hline
    \hline
    1 & \textbf{0.0020} & 0.0039 & \textbf{0.0020} & \textbf{0.0010} & 0.0046 & 0.193 \\
    2 & \textbf{0.0017} & 0.0048 & \textbf{0.0017} & \textbf{0.0012} & 0.0019 & 0.089 \\
    3 & 0.036 & 0.0042 & 0.036 & \textbf{0.0036} & \textbf{0.0020} & 0.146 \\
    4 & \textbf{0.0016} & 0.0057 & \textbf{0.0016} & \textbf{0.0021} & 0.0024 & 0.055 \\
    5 & 0.0021 & 0.0073 & 0.0021 & \textbf{0.0023} & \textbf{0.0020} & 0.044 \\
    6 & \textbf{0.0017} & 0.0107 & \textbf{0.0017} & \textbf{0.0010} & 0.0043 & 0.418 \\
    7 & \textbf{0.0015} & 0.0045 & \textbf{0.0015} & \textbf{0.0021} & 0.0017 & 0.075 \\
    8 & \textbf{0.0018} & 0.0079 & \textbf{0.0018} & \textbf{0.0069} & \textbf{0.0018} & 0.048 \\
    9 & \textbf{0.0019} & 0.0073 & \textbf{0.0019} & \textbf{0.0004} & \textbf{0.0019} & 0.113 \\
    \hline
  \end{tabular}
\end{table*}

\subsection{Metrics}

We evaluate estimated trajectories using two standard error metrics: the \emph{relative pose error} (RPE) and the \emph{absolute trajectory error} (ATE). 
RPE measures the local drift between consecutive poses and provides insight into odometry consistency, while ATE captures the global discrepancy between estimated and ground-truth trajectories after alignment. 

Formally, let $\{\mathbf{T}_i\}_{i=1}^N$ denote the ground-truth trajectory and $\{\hat{\mathbf{T}}_i\}_{i=1}^N$ the estimated trajectory, with $\mathbf{T}_i, \hat{\mathbf{T}}_i \in SE(3)$. 
The RPE evaluates the drift over a fixed time interval $\Delta$:
\begin{equation}
e^{\text{RPE}}_{i,\Delta} = \mathrm{trans}\!\left( (\mathbf{T}_i^{-1}\mathbf{T}_{i+\Delta})^{-1} \, (\hat{\mathbf{T}}_i^{-1}\hat{\mathbf{T}}_{i+\Delta}) \right),
\end{equation}
where $\mathrm{trans}(\cdot)$ extracts the translational component.
The root mean square error (RMSE) of RPE is then
\begin{equation}
\mathrm{RPE}_{\text{RMSE}} = \sqrt{ \frac{1}{N-\Delta} \sum_{i=1}^{N-\Delta} \left\| e^{\text{RPE}}_{i,\Delta} \right\|^2 }.
\end{equation}

On the other hand, ATE is defined as over the whole trajectory as
\begin{equation}
e^{\text{ATE}}_i = \mathrm{trans}\!\left( \mathbf{T}_i^{-1} \, \mathbf{S} \hat{\mathbf{T}}_i \right),
\end{equation}
where $\mathbf{S} \in SE(3)$ is an alignment transformation (if necessary). 
The corresponding RMSE is
\begin{equation}
\mathrm{ATE}_{\text{RMSE}} = \sqrt{ \frac{1}{N} \sum_{i=1}^N \left\| e^{\text{ATE}}_i \right\|^2 }.
\end{equation}
To allow comparison across traverses of different lengths, ATE is occasionally also normalized by the length of the associated ground-truth trajectory. 
Both metrics are computed using the \texttt{evo} library~\cite{grupp2017evo}.




\subsection{LuSNAR}

Table~\ref{tab:lusnar_results} summarizes the performance of our methods compared to ORB-SLAM3 across the nine LuSNAR scenes. 
VO denotes our baseline stereo odometry pipeline, while DEM-SLAM augments this baseline with DEM anchoring factors. 
The ORB-SLAM3 numbers are taken from the LuSNAR paper in its stereo configuration, and ATE values are normalized by ground truth trajectory length.
Figure~\ref{fig:lusnar_trajs} shows the ground truth, VO, and DEM-SLAM trajectories on Scene 9 of the LuSNAR dataset.

Across most scenes, our VO baseline achieves lower RPE than ORB-SLAM3, indicating more consistent short-term odometry. 
The addition of DEM anchoring does not alter RPE, as expected, but consistently reduces ATE by providing global surface constraints. 
This effect is most pronounced in longer traverses and visually repetitive environments, where ORB-SLAM3 accumulates global drift. 
These results confirm that DEM anchoring complements robust VO by improving long-range consistency without degrading local accuracy.

We note that LuSNAR DEMs are derived from trajectory-aligned LiDAR and therefore represent an optimistic anchoring scenario; this motivates our separate DEM uncertainty analysis at the end of this section.

\subsection{S3LI}



\begin{figure*}[t]
    \centering
    \subfloat[]{\includegraphics[width=0.45\linewidth]{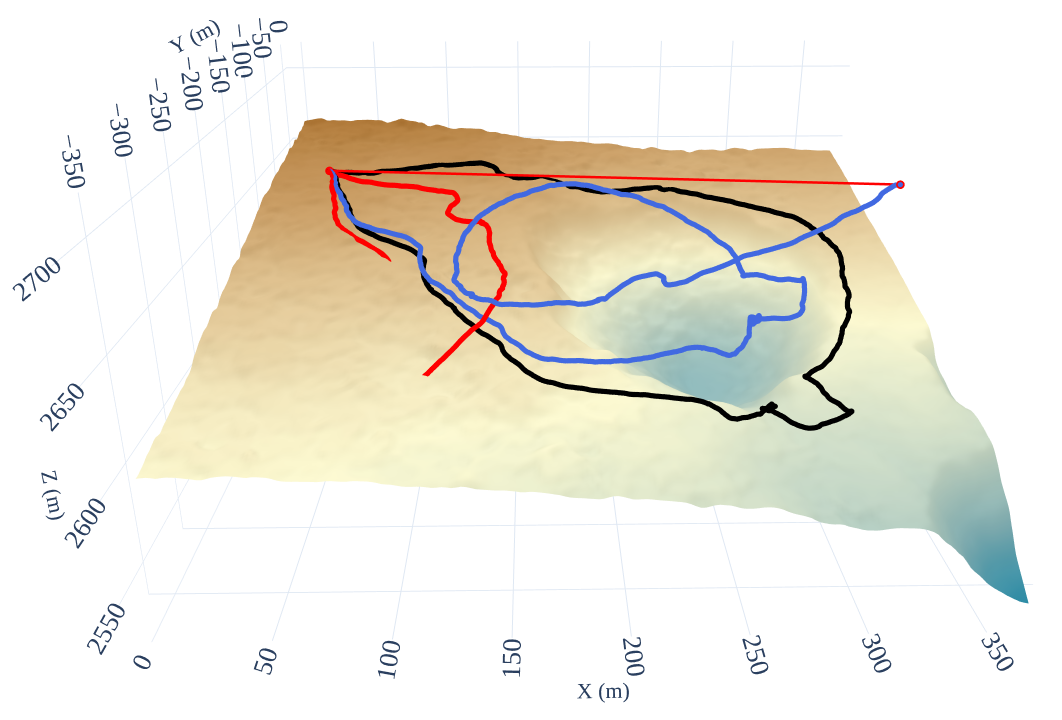}}\hfill
    \subfloat[]{\includegraphics[width=0.45\linewidth]{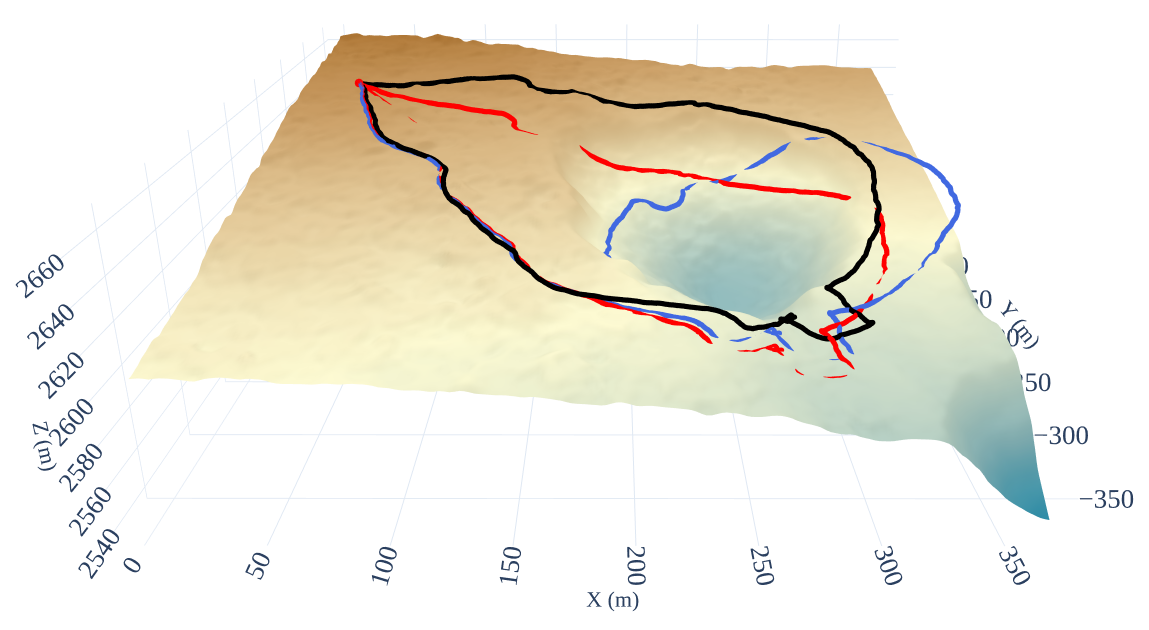}}\\
    \subfloat[]{\includegraphics[width=0.45\linewidth]{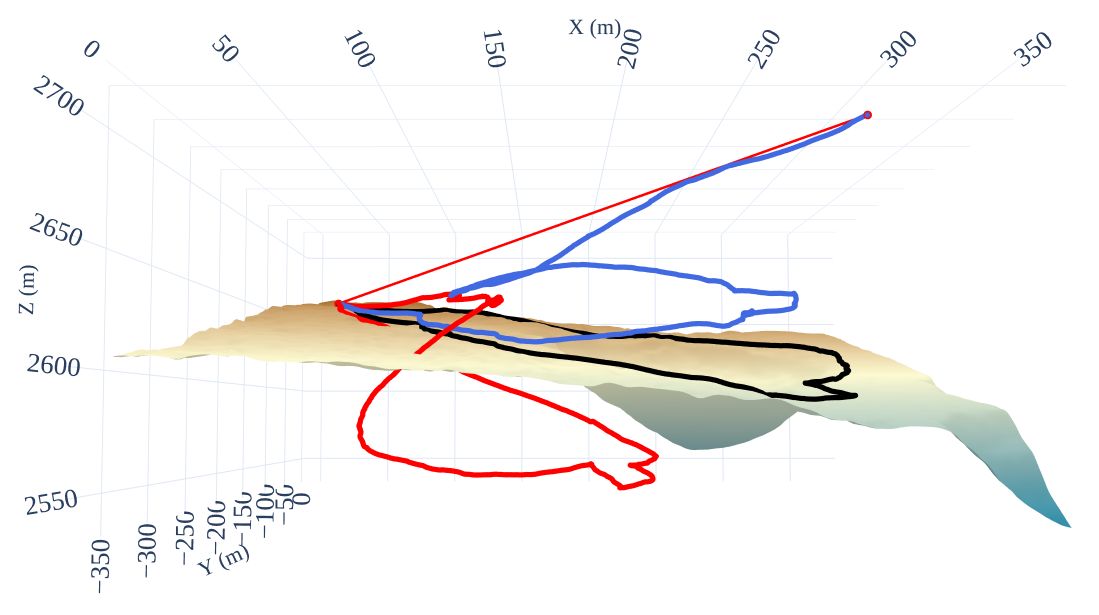}}\hfill
    \subfloat[]{\includegraphics[width=0.45\linewidth]{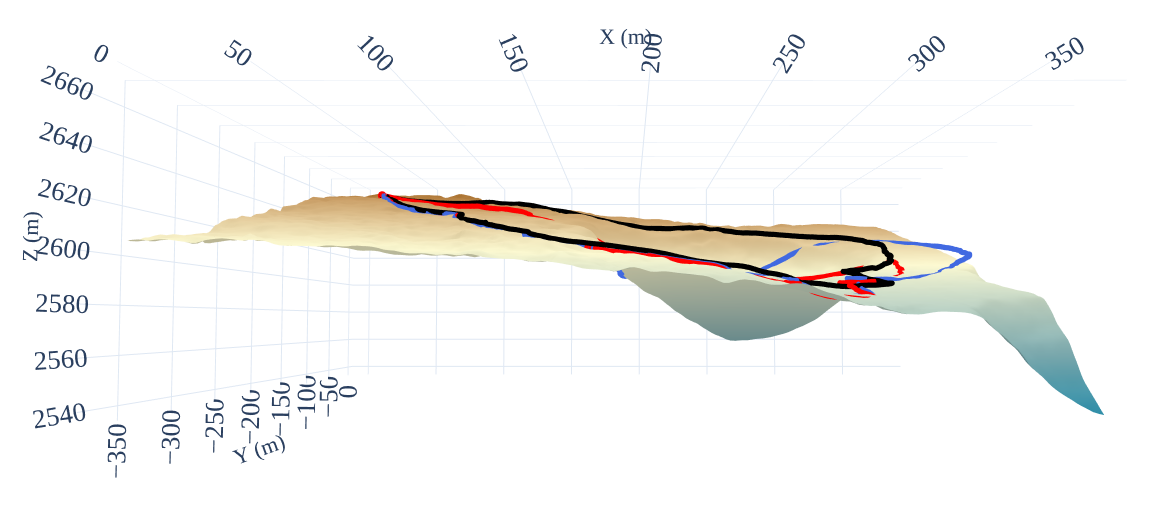}}
    \caption{S3LI crater loop with loop closure. Black: ground truth. Blue: before loop closure. Red: after loop closure. The loop closure connection (from start to end) is indicated with a straight red line. \textbf{Left}: loop closure without DEM anchoring leads to significant distortion. \textbf{Right}: DEM anchoring regularizes loop closure, mitigating drift and preventing catastrophic deformation. Residual deviation remains due to limited loop-closure constraints and the fact that DEM height and normal cues alone do not fully constrain tangential motion along visually aliased terrain.}
    \label{fig:s3li_crater_trajs}
\end{figure*}

We evaluate our approach on the \texttt{s3li\_crater} sequence, a 1.03 \unit{km} loop around the rim of the Cisternazza crater, which exhibits severe visual aliasing and minimal opportunities for loop closure. 
In this sequence, proximity-based loop closure is only possible near the start and end of the traverse.

Figure~\ref{fig:s3li_crater_trajs} compares loop closure behavior with and without DEM anchoring. 
Without DEM constraints, adding a loop closure factor successfully reconnects the trajectory but introduces significant distortion, pulling large portions of the estimated path away from the ground-truth rim and into the crater interior. 
This highlights that, in the presence of substantial accumulated drift, loop closure alone may not be sufficient to recover a physically plausible trajectory.

In contrast, when DEM anchoring is enabled, the loop-closed trajectory remains substantially closer to the ground-truth path. 
The DEM-derived height and surface-normal factors provide global surface context that regularizes the loop closure correction, preventing catastrophic deformation and mitigating long-term drift. 
While residual error remains and the reconstructed path does not perfectly recover the true rim traversal, DEM anchoring significantly improves global consistency relative to both visual odometry and loop-closed SLAM without DEM constraints.
After loop closure (and using the same ground-truth initial pose for both methods), the unconstrained trajectory exhibits an ATE RMSE of 94.01~\unit{m}, while the DEM-anchored trajectory achieves ATE RMSE of 21.43~\unit{m}.

These results suggest that DEM anchoring and loop closure play complementary roles: loop closure enforces global topological consistency, while DEM constraints help disambiguate geometrically plausible but physically incorrect solutions. 
Further improvements are expected through submap-to-DEM alignment and denser loop closure strategies, which we leave to future work.


\subsection{Unreal Engine Simulation}

\begin{figure*}[ht]
    \centering
    \begin{tabular}{cc}
    \includegraphics[width=0.49\linewidth]{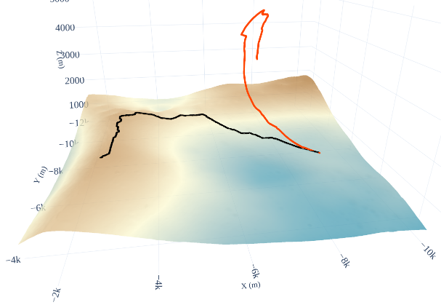} &
    \includegraphics[width=0.49\linewidth]{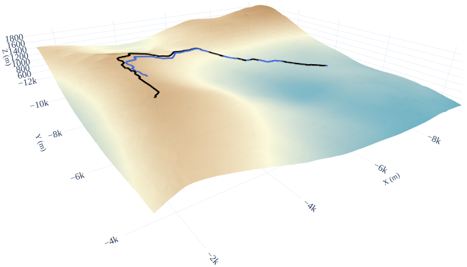} 
    \end{tabular}
    \caption{Long-range simulated traverse in Unreal Engine over an approximately 11 km trajectory. Ground truth is shown in black. \textbf{Left}: visual odometry (VO, orange) accumulates severe global drift over the traverse, deviating significantly from the true path. \textbf{Right}: DEM-anchored SLAM (blue) remains closely aligned with the ground truth but terminates early due to backend resource constraints, resulting in a shorter reconstructed trajectory while still exhibiting substantially reduced drift over the estimated portion.}
    \label{fig:unreal_trajs}
\end{figure*}

We further evaluate the proposed DEM-anchored SLAM system in a photorealistic Unreal Engine simulation constructed from LOLA-derived lunar DEMs, as described in Section~\ref{sec:datasets}. 
This environment enables controlled long-range traverses over realistic lunar topography while providing full ground truth for quantitative analysis.

Figure~\ref{fig:unreal_trajs} shows a representative long-range simulated trajectory of 11~\unit{km} in total length. 
The visual odometry (VO) baseline accumulates substantial drift over the traverse, deviating significantly from the ground-truth path. 
In contrast, the DEM-anchored SLAM trajectory remains well aligned with both the ground truth and the underlying terrain surface throughout the traverse. 
DEM anchoring effectively constrains global drift while preserving the local consistency of the visual odometry frontend.

Across multiple simulated traverses with injected pose noise, DEM anchoring consistently improves global accuracy without degrading local odometry performance. 
For a representative 1.19~\unit{km} trajectory, the VO baseline achieves an absolute trajectory error (ATE) of 78.0~\unit{m} and a relative pose error (RPE) of 0.017~\unit{m}. 
With DEM anchoring enabled, ATE is reduced to 7.71~\unit{m}—an order-of-magnitude improvement—while RPE remains essentially unchanged at 0.020~\unit{m}. 
These results demonstrate that DEM-derived surface constraints provide strong global regularization over long distances while maintaining accurate short-term motion estimates.


\subsection{DEM Uncertainty Analysis}
\label{subsec:dem_uncertainty}

\begin{table*}[t]
  \centering
  \caption{\bfseries DEM uncertainty sensitivity for DEM-anchored SLAM on a representative simulated traverse. Lower is better.}
  \label{tab:dem_uncertainty}
  \renewcommand{\arraystretch}{1.2}
  \begin{tabular}{|l|c|cc|}
    \hline
    \textbf{Perturbation} & \textbf{Level} & \textbf{ATE [m] $\downarrow$} & \textbf{Drift [\%] $\downarrow$} \\
    \hline
    \hline
    No DEM anchoring & -- & 0.788 & 0.986 \\
    DEM anchoring & -- & \textbf{0.672} & \textbf{0.717} \\
    \hline
    \hline
    Vertical noise (corr.) & 0.5 m RMS & 0.658 & 0.677 \\
    Vertical noise (corr.) & 1.0 m RMS & 0.894 & 0.898 \\
    Vertical noise (corr.) & 2.0 m RMS & 0.671 & 0.536 \\
    \hline
    Vertical bias & +0.5 m & 0.680 & 0.750 \\
    Vertical bias & +1.0 m & 0.712 & 0.801 \\
    \hline
    Horizontal shift & 0.5 m & 0.698 & 0.741 \\
    Horizontal shift & 1.0 m & 0.723 & 0.770 \\
    Horizontal shift & 2.0 m & 0.837 & 0.927 \\
    \hline
  \end{tabular}
\end{table*}

To evaluate robustness to realistic elevation model errors, we analyze the impact of DEM uncertainty on localization performance by injecting controlled perturbations into the reference DEM. We consider three common sources of DEM error: (i) spatially correlated vertical noise, (ii) constant vertical bias, and (iii) horizontal misregistration. These perturbations are designed to emulate typical error modes observed in orbital and photogrammetric DEM products.

Spatially correlated vertical noise is generated by sampling a smooth Gaussian random field and adding it to the DEM height values, with root-mean-square (RMS) amplitudes ranging from 0.5~\unit{m} to 2.0~\unit{m}. Constant vertical bias is modeled by applying a uniform height offset to the DEM, while horizontal misregistration is simulated by shifting the DEM sampling coordinates prior to querying surface height and surface normals. Unless otherwise noted, we fix the correlation length to 10~\unit{m}, apply horizontal shifts along the x-axis, and use a fixed random seed for reproducibility.
For this experiment we use a shorter traverse of length 378~\unit{m}, instead of the 11~\unit{km} traverse in Figure~\ref{fig:unreal_trajs}, to expedite analysis.

Table~\ref{tab:dem_uncertainty} summarizes localization performance under increasing DEM uncertainty for a representative simulated traverse. DEM anchoring consistently improves global accuracy relative to visual odometry across moderate levels of vertical noise and bias. As DEM error increases, performance degrades gracefully, with horizontal misregistration exhibiting the most pronounced impact due to its effect on both height and surface-normal consistency. Importantly, DEM anchoring does not degrade performance relative to visual odometry even under severe perturbations, indicating that the proposed DEM factors act as a stabilizing global prior rather than a brittle constraint.

Across all perturbations, relative pose error (RPE) remains unchanged, confirming that DEM anchoring primarily influences long-term global consistency without degrading local odometry accuracy.

These trends are consistent with prior analyses of lunar orbital DEM quality. As shown in~\cite{barker2021lola}, DEM errors are spatially structured and dominated by interpolation artifacts and sampling gaps, with typical RMS height errors of 0.3–0.5~\unit{m} and RMS slope errors of 1.5–2.5 degrees. In the Unreal Engine simulation, terrain geometry is generated directly from a LOLA DEM; the injected perturbations therefore represent plausible DEM–environment mismatches that may arise in real deployments.
\section{Conclusions}
\label{sec:conclusion}

In this work, we presented a stereo visual SLAM system for long-range lunar navigation that integrates learned feature matching with constraints derived from digital elevation models (DEMs). 
Our approach builds on a robust visual odometry baseline and introduces DEM-based height and surface-normal factors into the pose graph, providing absolute surface anchoring that mitigates drift over extended traverses.

We evaluated the system on three complementary datasets: the synthetic LuSNAR benchmark, the S3LI analog dataset based at Mt. Etna, and a custom Unreal Engine simulation environment built from LOLA South Pole DEMs. 
Across all datasets, DEM anchoring consistently reduced absolute trajectory error while leaving relative pose accuracy unchanged, confirming that local odometry remains reliable while global consistency is significantly improved. 
On challenging long-range traverses, such as the 1.03~\unit{km} crater sequence in S3LI, our method demonstrates improved accuracy and trajectory completion relative to prior visual and visual-inertial baselines.

These results highlight the potential of DEM anchoring as a scalable solution for future lunar missions, where GNSS is unavailable and repetitive terrain exacerbates drift in conventional pipelines. 
By fusing locally consistent stereo SLAM with globally referenced DEM constraints, our method enables reduced-drift, long-range navigation in unstructured planetary environments. 
Future work will extend this framework to multi-agent scenarios and integrate additional sensing modalities, further enhancing the autonomy and robustness required for sustained surface operations.


\acknowledgements 
This manuscript benefited from the use of AI-based assistants, which were employed for coding assistance and revisions. All final content was reviewed and edited by the authors.
The authors thank Blue Origin for funding this project and for valuable discussions that contributed to its development, and thank Daniel Neamati for reviewing this paper.

\bibliographystyle{IEEEtran}
\bibliography{references}

@article{campos2021orbslam3,
  title={{ORB-SLAM3: An Accurate Open-Source Library for Visual, Visual--Inertial, and Multi-Map SLAM}},
  author={Campos, Carlos and Elvira, Richard and Rodr{\'\i}guez, Juan J G{\'o}mez and Montiel, Jos{\'e} MM and Tard{\'o}s, Juan D},
  journal={IEEE transactions on robotics},
  volume={37},
  number={6},
  pages={1874--1890},
  year={2021},
  publisher={IEEE}
}

@inproceedings{detone_superpoint_2018-1,
  location   = {Salt Lake City, {UT}, {USA}},
  title      = {{SuperPoint: Self-Supervised Interest Point Detection and Description}},
  isbn       = {978-1-5386-6100-0},
  url        = {https://ieeexplore.ieee.org/document/8575521/},
  doi        = {10.1109/CVPRW.2018.00060},
  shorttitle = {{SuperPoint}},
  eventtitle = {2018 {IEEE}/{CVF} Conference on Computer Vision and Pattern Recognition Workshops ({CVPRW})},
  pages      = {337--33712},
  booktitle  = {2018 {IEEE}/{CVF} Conference on Computer Vision and Pattern Recognition Workshops ({CVPRW})},
  publisher  = {{IEEE}},
  author     = {{DeTone}, Daniel and Malisiewicz, Tomasz and Rabinovich, Andrew},
  urldate    = {2025-06-09},
  date       = {2018-06},
  year       = 2018,
  langid     = {english}
}

@inproceedings{lindenberger2023lightglue,
  title={{LightGlue: Local Feature Matching at Light Speed}},
  author={Lindenberger, Philipp and Sarlin, Paul-Edouard and Pollefeys, Marc},
  booktitle={Proceedings of the IEEE/CVF International Conference on Computer Vision},
  pages={17627--17638},
  year={2023}
}

@misc{gtsam,
  author       = {Frank Dellaert and GTSAM Contributors},
  title        = {borglab/gtsam},
  month        = May,
  year         = 2022,
  publisher    = {Georgia Tech Borg Lab},
  version      = {4.2a8},
  doi          = {10.5281/zenodo.5794541},
  url          = {https://github.com/borglab/gtsam)}
}

@article{maimone2007two,
  title={{Two Years of Visual Odometry on the Mars Exploration Rovers}},
  author={Maimone, Mark and Cheng, Yang and Matthies, Larry},
  journal={Journal of Field Robotics},
  volume={24},
  number={3},
  pages={169--186},
  year={2007},
  publisher={Wiley Online Library}
}

@article{verma2023autonomous,
  title={{Autonomous robotics is driving Perseverance rover’s progress on Mars}},
  author={Verma, Vandi and Maimone, Mark W and Gaines, Daniel M and Francis, Raymond and Estlin, Tara A and Kuhn, Stephen R and Rabideau, Gregg R and Chien, Steve A and McHenry, Michael M and Graser, Evan J and others},
  journal={Science Robotics},
  volume={8},
  number={80},
  pages={eadi3099},
  year={2023},
  publisher={American Association for the Advancement of Science}
}

@inproceedings{daftry2023lunarnav,
  title={{LunarNav: Crater-based localization for long-range autonomous lunar rover navigation}},
  author={Daftry, Shreyansh and Chen, Zhanlin and Cheng, Yang and Tepsuporn, Scott and Khattak, Shehryar and Matthies, Larry and Coltin, Brian and Naal, Ussama and Ma, Lanssie Mingyue and Deans, Matthew},
  booktitle={2023 IEEE Aerospace Conference},
  pages={1--15},
  year={2023},
  organization={IEEE}
}

@article{atha2024shadownav,
  title={{ShadowNav: Autonomous Global Localization for Lunar Navigation in Darkness}},
  author={Atha, Deegan and Swan, R Michael and Cauligi, Abhishek and Bettens, Anne and Goh, Edwin and Kogan, Dima and Matthies, Larry and Ono, Masahiro},
  journal={IEEE Transactions on Field Robotics},
  year={2024},
  publisher={IEEE}
}

@inproceedings{nefian2014planetary,
  title={{Planetary Rover Localization within Orbital Maps}},
  author={Nefian, Ara V and Bouyssounouse, Xavier and Edwards, L and Kim, Taemin and Hand, E and Rhizor, Jared and Deans, M and Bebis, George and Fong, Terrence},
  booktitle={2014 IEEE International Conference on Image Processing (ICIP)},
  pages={1628--1632},
  year={2014},
  organization={IEEE}
}

@article{zhang2024lidar,
  title={{LiDAR-Inertial SLAM with DEM-driven Global Constraints for Planetary Rover Exploration}},
  author={Zhang, Xusheng and Li, Yuan and Cao, Zeyuan and Lv, Junying and Huang, Zefeng and Zhang, Wuming},
  journal={The International Archives of the Photogrammetry, Remote Sensing and Spatial Information Sciences},
  volume={48},
  pages={615--621},
  year={2024},
  publisher={Copernicus GmbH}
}

@article{melman2022lcns,
  title={{LCNS Positioning of a Lunar Surface Rover Using a DEM-based Altitude Constraint}},
  author={Melman, Floor Thomas and Zoccarato, Paolo and Orgel, Csilla and Swinden, Richard and Giordano, Pietro and Ventura-Traveset, Javier},
  journal={Remote Sensing},
  volume={14},
  number={16},
  pages={3942},
  year={2022},
  publisher={MDPI}
}

@article{barker2021lola,
  title={{Improved LOLA Elevation Maps for South Pole Landing Sites: Error Estimates and Their Impact on Illumination Conditions}},
  author={Barker, Michael K and Mazarico, Erwan and Neumann, Gregory A and Smith, David E and Zuber, Maria T and Head, James W},
  journal={Planetary and Space Science},
  volume={203},
  pages={105119},
  year={2021},
  publisher={Elsevier}
}

@article{furgale2012devon,
  title={The Devon Island rover navigation dataset},
  author={Furgale, Paul and Carle, Pat and Enright, John and Barfoot, Timothy D},
  journal={The International Journal of Robotics Research},
  volume={31},
  number={6},
  pages={707--713},
  year={2012},
  publisher={SAGE Publications Sage UK: London, England}
}

@inproceedings{vayugundla2018datasets,
  title={{Datasets of Long Range Navigation Experiments in a Moon Analogue Environment on Mount Etna}},
  author={Vayugundla, Mallikarjuna and Steidle, Florian and Smisek, Michal and Schuster, Martin J and Bussmann, Kristin and Wedler, Armin},
  booktitle={ISR 2018; 50th International Symposium on Robotics},
  pages={1--7},
  year={2018},
  organization={VDE}
}

@article{meyer2021madmax,
  title={{The MADMAX data set for visual-inertial rover navigation on Mars}},
  author={Meyer, Lukas and Sm{\'\i}{\v{s}}ek, Michal and Fontan Villacampa, Alejandro and Oliva Maza, Laura and Medina, Daniel and Schuster, Martin J and Steidle, Florian and Vayugundla, Mallikarjuna and M{\"u}ller, Marcus G and Rebele, Bernhard and others},
  journal={Journal of Field Robotics},
  volume={38},
  number={6},
  pages={833--853},
  year={2021},
  publisher={Wiley Online Library}
}

@inproceedings{Dai2025LAC,
    author    = {Adam Dai and Asta Wu and Keidai Iiyama and Guillem Casadesus Vila and Kaila Coimbra and Thomas Deng and Grace Gao},
    title     = {{Full Stack Navigation, Mapping, and Planning for the Lunar Autonomy Challenge}},
    booktitle = {Proceedings of ION GNSS+ 2025},
    address   = {Baltimore, MD, USA},
    month     = sep,
    year      = {2025},
    publisher = {The Institute of Navigation}
}

@article{liu2024lusnar,
  title={{LuSNAR: A Lunar Segmentation, Navigation and Reconstruction Dataset based on Muti-sensor for Autonomous Exploration}},
  author={Liu, Jiayi and Zhang, Qianyu and Wan, Xue and Zhang, Shengyang and Tian, Yaolin and Han, Haodong and Zhao, Yutao and Liu, Baichuan and Zhao, Zeyuan and Luo, Xubo},
  journal={arXiv preprint arXiv:2407.06512},
  year={2024}
}

@article{giubilato2022challenges,
  title={{Challenges of SLAM in Extremely Unstructured Environments: The DLR Planetary Stereo, Solid-State LiDAR, Inertial Dataset}},
  author={Giubilato, Riccardo and St{\"u}rzl, Wolfgang and Wedler, Armin and Triebel, Rudolph},
  journal={IEEE Robotics and Automation Letters},
  volume={7},
  number={4},
  pages={8721--8728},
  year={2022},
  publisher={IEEE}
}

@misc{grupp2017evo,
  title={{evo: Python package for the evaluation of odometry and SLAM.}},
  author={Grupp, Michael},
  howpublished={\url{https://github.com/MichaelGrupp/evo}},
  year={2017}
}

@article{palaseanu2020digital,
  title={{Digital Surface Model of Mt. Etna, Italy, derived from 2015 Pleiades Satellite Imagery}},
  author={Palaseanu-Lovejoy, M and Bisson, M and Spinetti, C and Buongiorno, MF and Alexandrov, O and Cerere, T},
  journal={US Geological Survey data release, https://doi. org/10.5066/P9IGLDYE},
  year={2020}
}

@article{qin2018vins,
  title={{VINS-Mono: A Robust and Versatile Monocular Visual-Inertial State Estimator}},
  author={Qin, Tong and Li, Peiliang and Shen, Shaojie},
  journal={IEEE transactions on robotics},
  volume={34},
  number={4},
  pages={1004--1020},
  year={2018},
  publisher={IEEE}
}

@article{zhao2025light,
  title={{Light-SLAM: A Robust Deep-Learning Visual SLAM System Based on LightGlue under Challenging Lighting Conditions}},
  author={Zhao, Zhiqi and Wu, Chang and Kong, Xiaotong and Li, Qiyan and Guo, Zifan and Lv, Zejie and Du, Xiaoqi},
  journal={IEEE Transactions on Intelligent Transportation Systems},
  year={2025},
  publisher={IEEE}
}

@article{revaud2019r2d2,
  title={{R2D2: Reliable and Repeatable Detector and Descriptor}},
  author={Revaud, Jerome and De Souza, Cesar and Humenberger, Martin and Weinzaepfel, Philippe},
  journal={Advances in neural information processing systems},
  volume={32},
  year={2019}
}

@techreport{merancy2024moon,
    author = {Merancy, Nujoud F},
    title = {{Moon to Mars Architecture Executive Overview}},
    institution = {National Aeronautics and Space Administration},
    year = {2024}
}

@misc{opportunity_rover_asset,
  title        = {{Mars Exploration Rover Simulator}},
  author       = {{Fab Marketplace}},
  howpublished = {\url{https://www.fab.com/listings/97365511-4f76-41ac-8e1c-ef0d4c75e5ad}},
  note         = {Accessed: Jan. 2026},
  year         = {2024}
}

@inproceedings{vila2025lupnt,
  title={{LuPNT: An Open-Source Simulator for Lunar Communications, Positioning, Navigation, and Timing}},
  author={Vila, Guillem Casadesus and Iiyama, Keidai and Gao, Grace},
  booktitle={2025 IEEE Aerospace Conference},
  pages={1--18},
  year={2025},
  organization={IEEE}
}




\thebiography
\begin{biographywithpic}
{Adam Dai}{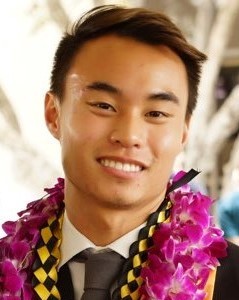}
is a Ph.D. candidate in the Department of Electrical Engineering at Stanford University. He received his B.Sc. degree in Electrical Engineering with a minor in Computer Science in 2019 from the California Institute of Technology. His research interests include navigation, mapping, and planning in unstructured 3D environments.
\end{biographywithpic} 

\begin{biographywithpic}
{Guillem Casadesus Vila}{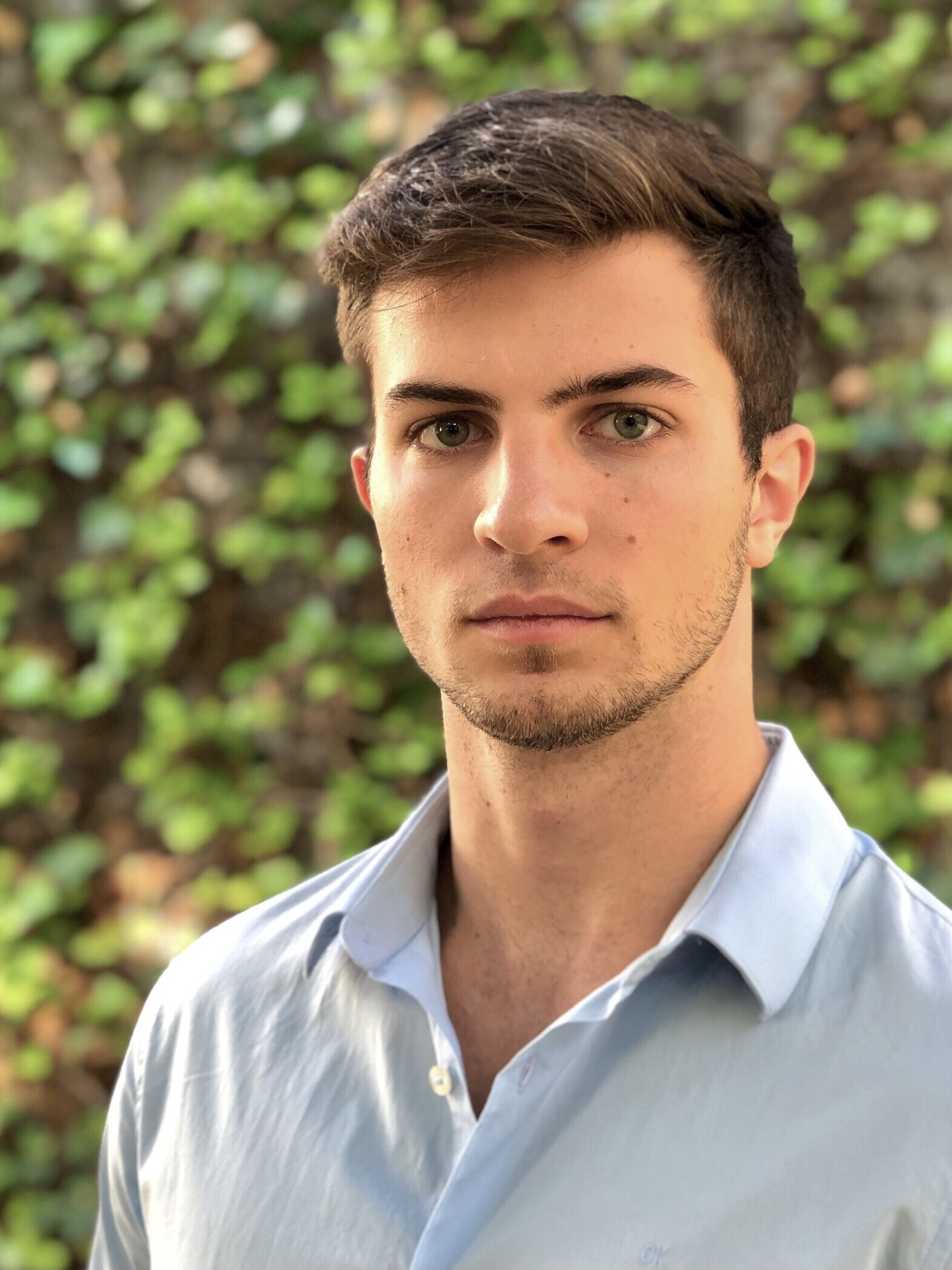}
is a Ph.D. candidate in the Department of Aeronautics and Astronautics at Stanford University. He received his B.Sc. degree in Aerospace and Telecommunications Engineering in 2022 from the Universitat Politecnica de Catalunya (UPC) under the CFIS program. His research interests include robotic space exploration, autonomous and distributed space systems, as well as navigation and communication satellites.
\end{biographywithpic} 

\begin{biographywithpic}
{Grace Gao}{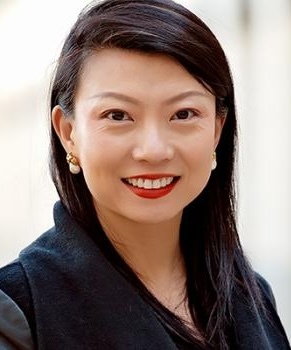}
is an associate professor in the Department of Aeronautics and Astronautics at Stanford University.
Before joining Stanford University, she was an assistant professor at University of Illinois at Urbana-Champaign.
She obtained her Ph.D. degree at Stanford University.
Her research is on robust and secure positioning, navigation, and timing with applications to manned and unmanned aerial vehicles, autonomous driving cars, as well as space robotics.
\end{biographywithpic}
\vfill

\end{document}